# Aircraft Detection in Satellite Imagery using Deep Learning Object Detectors

Mujahir Hussain Abbasi
*Department of Information Technology*
*Maulana Abul Kalam Azad University of Technology,*
Kolkata, West Bengal, India.
mujahirabbasi@gmail.com

S. Beghin Bose
*School of Computational and Physical Sciences*
*Kristu Jayanti (Deemed to be University)*
K. Narayanapura, Kothanur Post,
Bengaluru - 560077, Karnataka, India.
s.beghinbose@gmail.com

A. T. R. Krishna Priya
*AB Technologies*
Nagercoil, Kanyakumari,
Tamil Nadu, India.
priya.gita6@gmail.com

P. Jamuna
*Department of Electrical and Electronics Engineering*
*Nandha Engineering College,*
Erode-638052. India.
ravikumarjamuna@gmail.com

G. Anandhakumari
*Department of Physics*
*Erode Sengunthar Engineering College,*
Perundurai -638057, India.
anandhipriya95@gmail.com

S. Jaanaa Rubavathy
*Department of Electrical and Electronics Engineering*
*Saveetha School of Engineering, Saveetha Institute of Medical and Technical Sciences (SIMATS),*
*Saveetha University,*
Chennai, India.
jaanaaruba@gmail.com

***Abstract*—The object detection in satellite imagery has garnered considerable attention due to its extensive real-world applications and the inherent challenges it presents, including noise, fluctuating image quality, and intricate backgrounds. This paper proposed a framework for object detection that combines image enhancement and Deep Learning (DL) to make detection more accurate. First, a Gabor filter is used to process the input image to bring out important features and reduce noise. Then, normalization is used to make sure that the data is evenly distributed so that the model work properly. After that, a model based on YOLOv11 is used to quickly learn and find object features. The proposed method has a high mAP of 95%, a precision of 97%, a recall of 85%, and an F1-score of 91%, showing that it works better than others. Finally, the system makes predictions by correctly identifying objects and putting bounding boxes around areas that have been found. This makes it useful for real-time applications.**

## I. Introduction

Aircraft have become increasingly important since the beginning of economic globalization, which has led to a greater focus on aircraft as part of the aviation industry and the need for an accurate object detection system for aircraft [1-2]. While it is relatively easy to detect an aircraft using a traditional terrestrial object detection system at an airport, there are significant differences between the detected object and the airport background with respect to ambient lighting. Because the background and object do not have a uniform or consistent distribution, there also be a considerable variance when attempting to detect the object using a remote sensing imaging system (RSI) [3-4]. For small-scale objects, this makes locating the object very difficult.

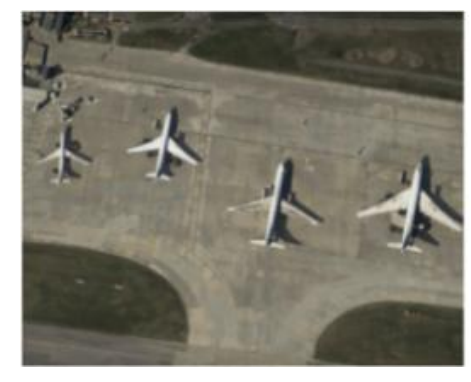
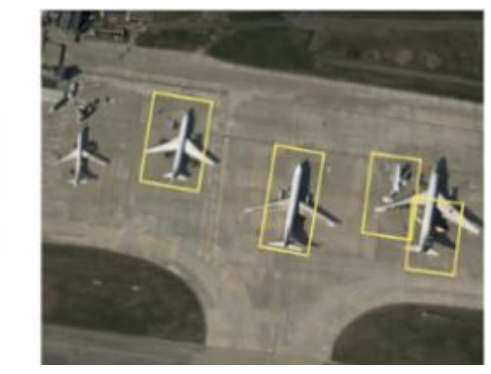

Fig. 1. Aircraft Detection in Satellite

The Aircraft Detection and Classification has important real world benefits and play a major role in many different fields including Military Reconnaissance, Military Decision-Making, Air Traffic Control, Autonomous Flight, and Automatic Navigation. There are many challenges associated with aircraft detection and classification in remote sensing through satellite imagery [5-6]. First, the resolution of satellite generated remote sensing imagery is low therefore, an aircraft target represented by only 10 to 20 pixels when viewed in a satellite image resulting in difficulty in targeting an aircraft [7-8]. Second, providing additional challenges, aircraft target located in or beneath cloud cover, on mountain tops or in settings where there is snow, therefore requiring additional efforts to identify and classify an aircraft target. In addition to aircraft having different heights, sizes, and orientations because remote sensing satellites shoot them at different angles, the models used for detecting these aircraft should be able to generalize to multiple different possible ways. Furthermore, due to cloud or building occlusions, the aircraft target difficult to detect completely [9-10]. Lastly, because of the class imbalance (due to size of some classes) and sensitivity of different classes, some classes have a higher chance of being represented by more than their share, thereby introducing more uncertainty while training and detecting these targets [11-12]. An efficient pre-processing and classification method is proposed to improve image quality and increase the precision of aircraft detection to overcome these difficulties.

The number of lightweight DL object detection models are examined to address the difficulties in detecting aircraft from satellite imagery. These models were chosen because of their capacity to strike a balance between computational efficiency and detection accuracy. YOLO-based nano and tiny variants are compared to assess the way they perform in limited conditions.

TABLE I. COMPARISON ANALYSIS OF LIGHTWEIGHT YOLO BASED OBJECT DETECTION MODELS FOR AIRCRAFT DETECTION

| Method | Operations | Advantages | Drawbacks |
|---|---|---|---|
| YOLOv5 [13] | YOLOv5 Nano employs a lightweight backbone with PANet for multi-scale feature fusion and CSP for feature extraction. It detects objects quickly and with less complexity in a single stage. | Fast inference is made possible by its small model size and extremely low computational cost. It is therefore appropriate for embedded and real-time applications. | Because of its simplified architecture, it offers lower detection accuracy. Additionally, it has trouble identifying extremely small objects in satellite imagery. |
| YOLOv7 Tiny [14] | For improved feature representation, YOLOv7 Tiny employs an optimized backbone with effective layer aggregation. In a single-stage framework, it facilitates multi-scale object detection. | When compared to YOLOv5 Nano, it provides better accuracy and feature extraction. It keeps efficiency and speed in a healthy balance. | Compared to lighter models, it needs a little more processing power. In extremely complicated detection scenarios, its performance is still restricted. |
| YOLOv8 Nano [15] | YOLOv8 Nano employs an enhanced feature pyramid network and an anchor-free detection method. It facilitates effective multi-scale detection through a streamlined training procedure. | Among nano models, it offers enhanced small object detection and greater accuracy. Additionally, it achieves improved generalization and faster convergence. | Compared to previous nano versions, it is a little more complex. For proper implementation, it also needs updated frameworks. |

An enhanced YOLOv11 model is proposed for better accuracy and reliable aircraft detection to get around these restrictions.

In summary, lightweight models exhibit good efficiency but are limited in complex scenarios. For dependable aircraft detection, the proposed YOLOv11 increases accuracy and robustness.

*A. Objectives*

- To improve the quality of the input image by applying a Gabor filter for efficient feature enhancement and noise reduction.
- To normalize image data so that model input is dependable and consistent.
- To use YOLOv11 to create and train an object detection model for precise feature learning.
- To assess the model's performance to guarantee detection accuracy.
- To precisely locate target regions by producing bounding boxes and accurately detecting objects.

## II. PROPOSED SYSTEM DESCRIPTION

An object detection framework with structured processing is depicted in the Fig. 2. A Gabor filter is first applied to the input image during a pre-processing step in to highlight significant features and minimize noise. After processing, the image is normalized by scaling the pixel values to provide consistent model input.

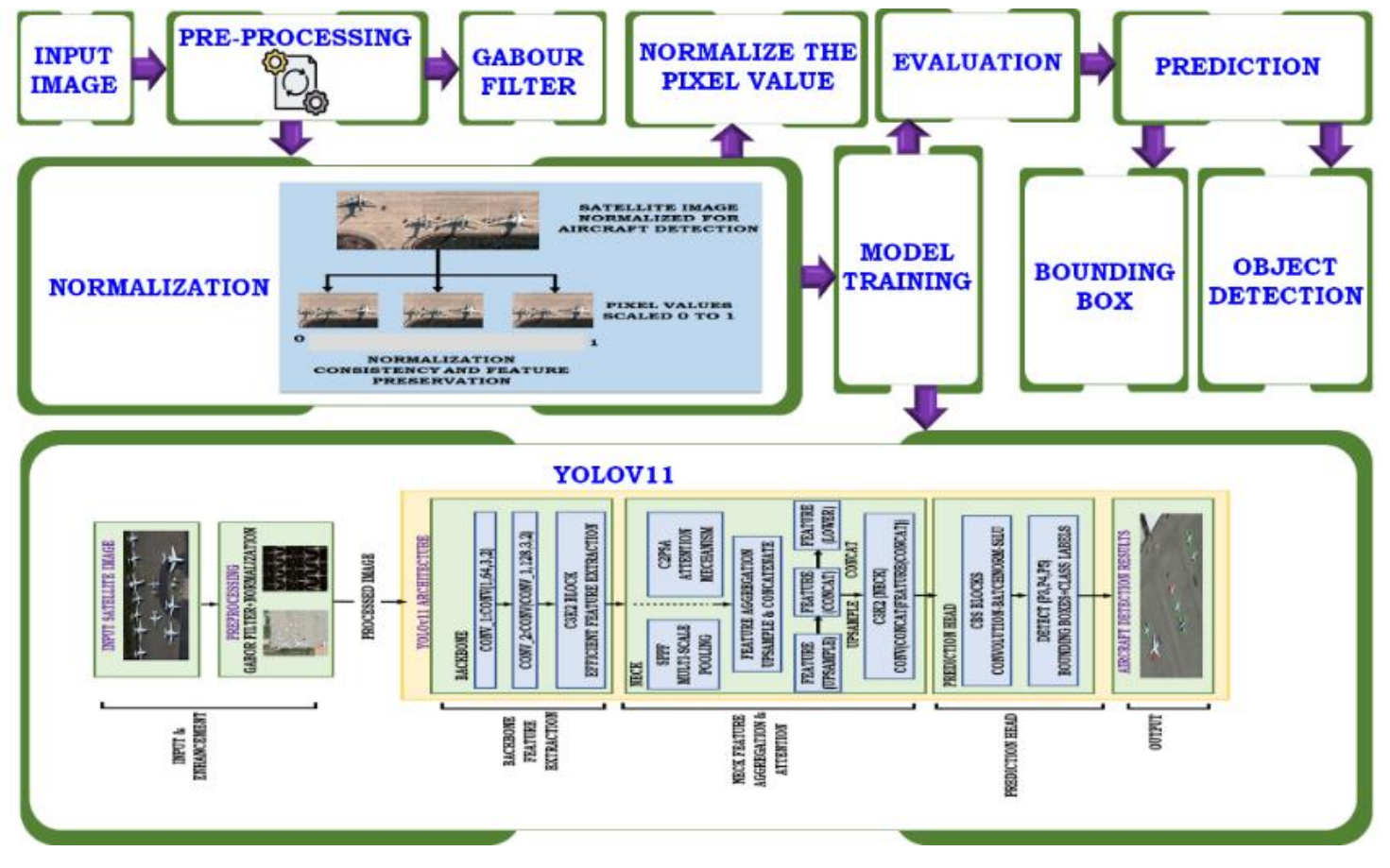

Fig. 2. Block diagram of the proposed work

The system then efficiently learn object features by feeding the normalized data into the model training phase using YOLOv11. After training, the model is assessed to make sure it performs accurately. To produce accurate object detection results, the system first makes a prediction that identifies objects and creates bounding boxes around detected regions.

## III. PROPOSED SYSTEM MODELLING

*A. Gabor Filter*

By accurately identifying aircraft becomes difficult in satellite imagery due to noise, fluctuating illumination, and complex backgrounds. A Gabor filter is used as an efficient image enhancement method to solve these problems because it capture spatial frequency and orientation information that is comparable to the human visual system. Because of this, it is especially useful for drawing attention to the edges, wings, and fuselage patterns of aircraft in satellite images.

A 2-D Gabor filter allows for simultaneous localization in both the spatial and frequency domains. It is defined as a Gaussian kernel modulated by a sinusoidal wave. In mathematical terms, it is represented as:

$$\Psi(x, y) = \exp - \frac{x'^2 + y'^2}{2\sigma^2} \cdot \exp(\omega x') \qquad (1)$$

Where (x, y) stands for pixel coordinates, $\omega$ is the central frequency, $\theta$ indicates the filter's orientation, and σ regulates the Gaussian function's spread. $x' = x\cos\theta + y\sin\theta$ and $y' = -x\sin\theta + y\cos\theta$ are used.

A Gabor filter bank with several frequencies and orientations typically five scales and eight orientations is used for efficient aircraft feature extraction. To create feature representations, the Gabor filters are convolved with the input satellite image I(x, y):

$$G_{m,n}(x,y) = I(x,y) * \Psi(w_m, \theta_n)(x,y) \quad (2)$$

The important aircraft features are effectively enhanced while background noise is suppressed by using the magnitude of the convolution output. Accurate aircraft detection in later processing stages is supported by this enhanced feature representation.

### *B. Normalization*

The improved satellite image is normalized to ensure consistent data representation for aircraft detection and to normalize the pixel value. It reduces variations brought on by illumination and sensor differences by scaling pixel intensities to a fixed range, usually between 0 and 1. Important aircraft features like edges and shapes are preserved by this process. Consequently, it enhances model performance and detection accuracy.

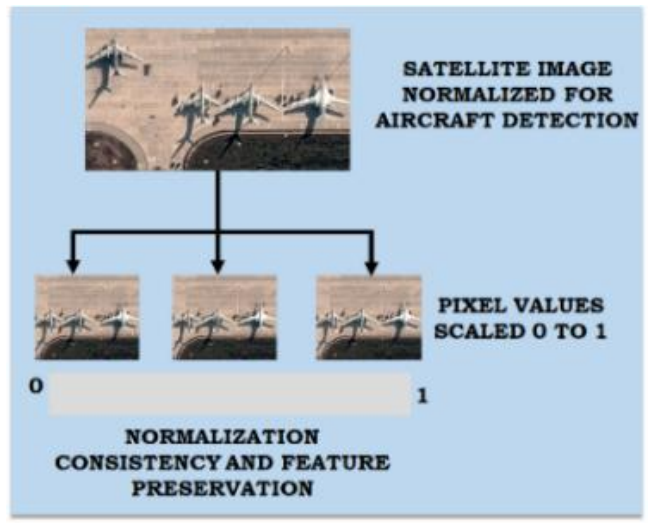

Fig. 3. Normalization

### *C. Classification*

The processed satellite images are fed into the YOLOv11 model for effective aircraft detection after image enhancement using the Gabor filter and normalization. The model's enhanced architectural elements enable precise aircraft identification in complicated backgrounds and at different resolutions.

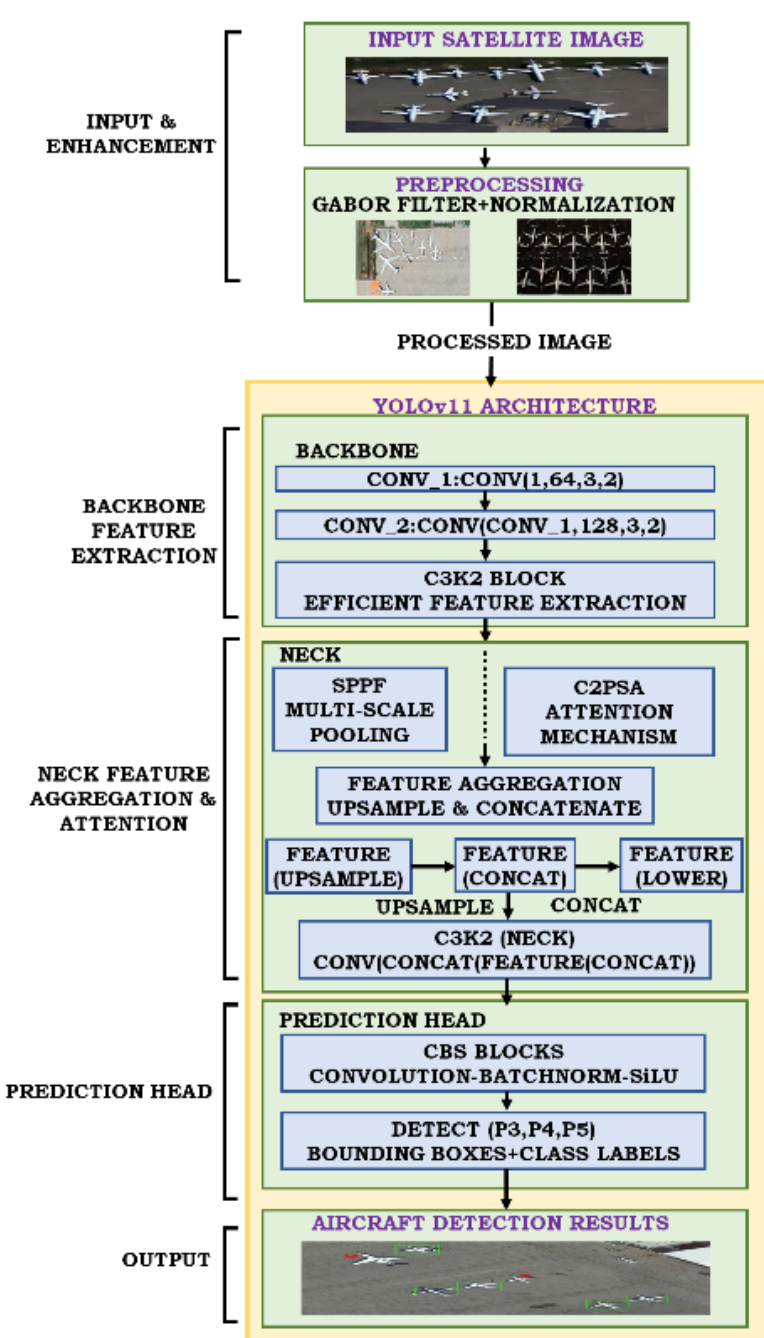

Fig. 4. Architectur of YOLOv11

YOLOv11's core makes use of an improved C3k2 block, which lowers computational complexity and allows for effective feature extraction. Convolutional layers are first used to down sample the input image to capture key spatial features:

$$Conv_1 = conv(1, 64, 3, 2) \quad (3)$$

$$Conv_2 = Conv(Conv_1, 128, 3, 2) \quad (4)$$

By dividing and combining feature maps, the C3k2 block enhances feature representation:

$$C3K2(X) = Conv\big(split(X)\big) + conv\Big(Merge\big(split(X)\big)\Big) \quad (5)$$

The neck of YOLOv11 integrates attention mechanisms through the C2PSA block, which highlights significant areas like aircraft in cluttered environments, to improve detection in complex satellite scenes:

$$C2PSA(X) = Attention(Concat)(X_1, x_2) \quad (6)$$

Furthermore, multi-scale feature extraction uses spatial pyramid pooling:

$$SPPF(X) = Concat\big(Maxpool(X,5), Maxpool(X,3), Maxpool(X,1)\big) \quad (7)$$

The neck additionally aggregates features by combining multi-scale data through concatenation and up sampling:

$$Feature(upsample) = Upsample\big(Feature(previous)\big) \quad (8)$$

$$Feature(concat) = concat(Feature(upsample), Feature(lower)\ ) \quad (9)$$

Convolution operations are used to refine these aggregated features:

$$C3K2(neck) = Conv\Big(Concat\big(Feature(concat)\big)\Big) \quad (10)$$

Ultimately, the prediction head uses CBS (Convolution–BatchNorm–SiLU) blocks to produce precise detection outputs, such as class labels and bounding box coordinates:

$$Detect(P3, P4, P5) = BoundingBoxes + ClassLabels \quad (11)$$

The Robust learning of aircraft features from satellite images is made possible by this architecture, even in the face of background complexity and noise. The model's performance is assessed, and it generates accurate predictions by precisely identifying aircraft and encircling them with bounding boxes, ensuring dependable and instantaneous object detection.

## IV. RESULT AND DISCUSSION

The results and discussion assess the proposed model utilizing the Aircraft Detection Computer Vision Dataset. It emphasizes performance, pre-processing, and accuracy of detection.

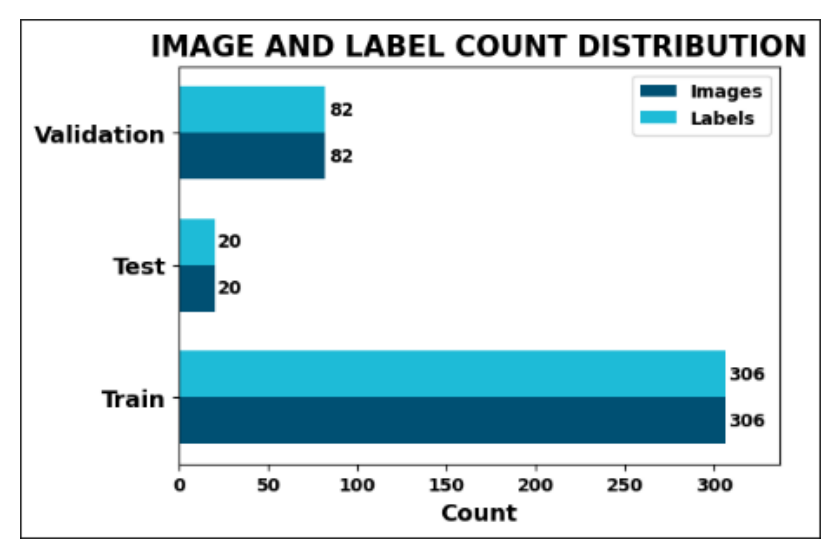

Fig. 5. Image and label count distribution across dataset splits

The images and labels in the training, testing, and validation data sets are shown in Fig. 5. The training set contains the largest number of samples at (306), followed by the validation set with (82) and the testing set with (20). Each split contains the same number of images to their respective labels.

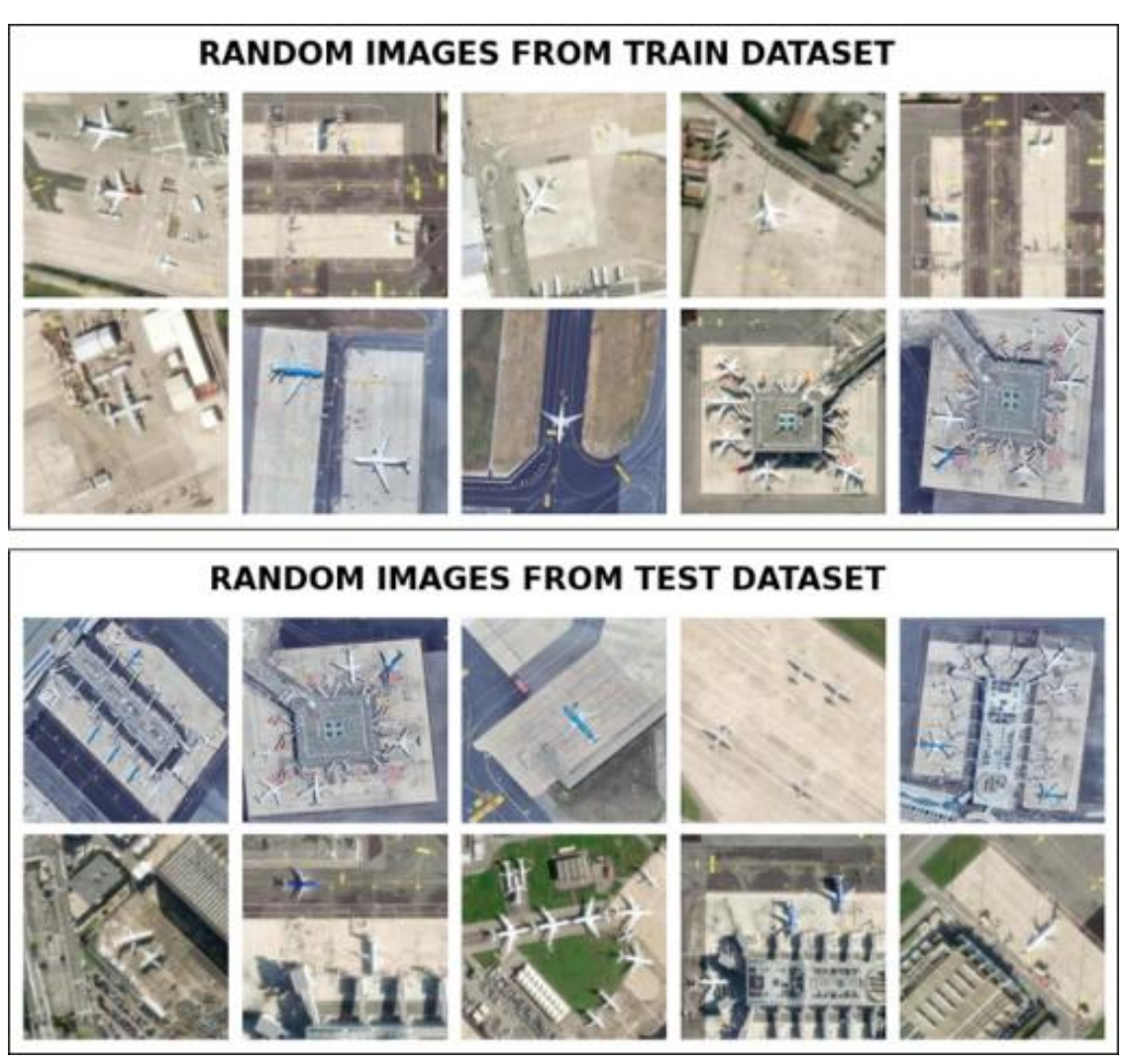

Fig. 6. Sample visualization of training and testing dataset images

The Random images from the training set and testing set are depicted in Fig. 6. Each of the images demonstrates that there are a variety of planes, rotated in different ways, and with diverse backgrounds. As a result, this demonstrates the complexity and the diversity of the dataset, which are both critical to assisting the model in becoming more universally applicable and to improving its ability to locate objects.

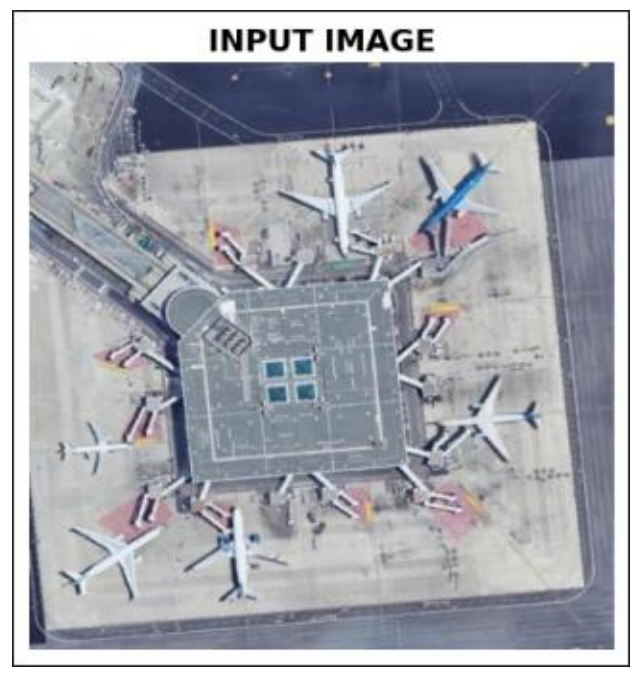

Fig. 7. Sample input image of airport scene

Figure 7 show an aerial shot of an airport where many planes have been parked near the terminal building(s) each lines up a different size and orientation relative to one another, as well as the many associated infrastructures that surround them, thus making this scene difficult for detecting objects by ML means.

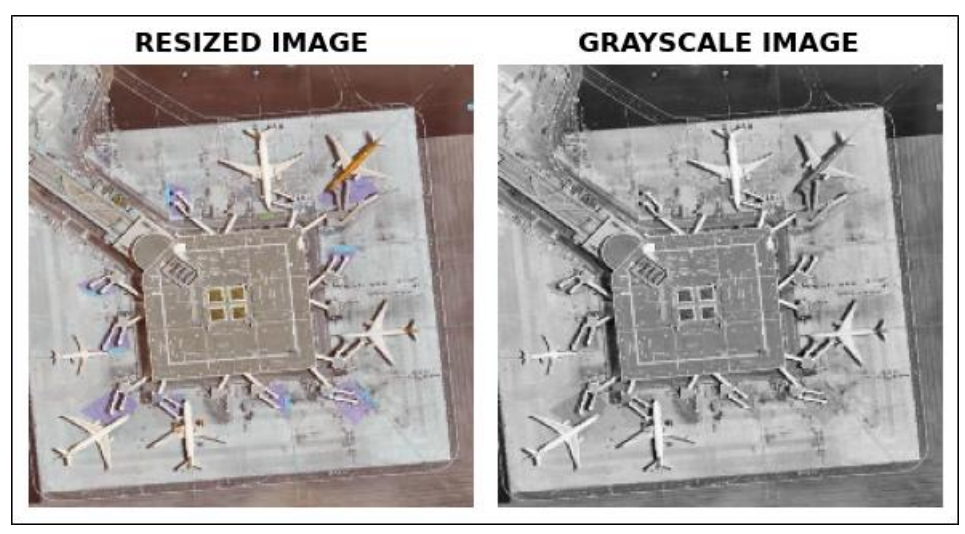

Fig. 8. Image pre-processing resizing and grayscale conversion

The Preparation of input image in Fig. 8 involved resizing the input image to a standard size and converting it to grayscale. By performing these two steps, the calculations to create the model using the prepared input image will be less complex, while retaining the important structural characteristics, to provide a good basis for training the model.

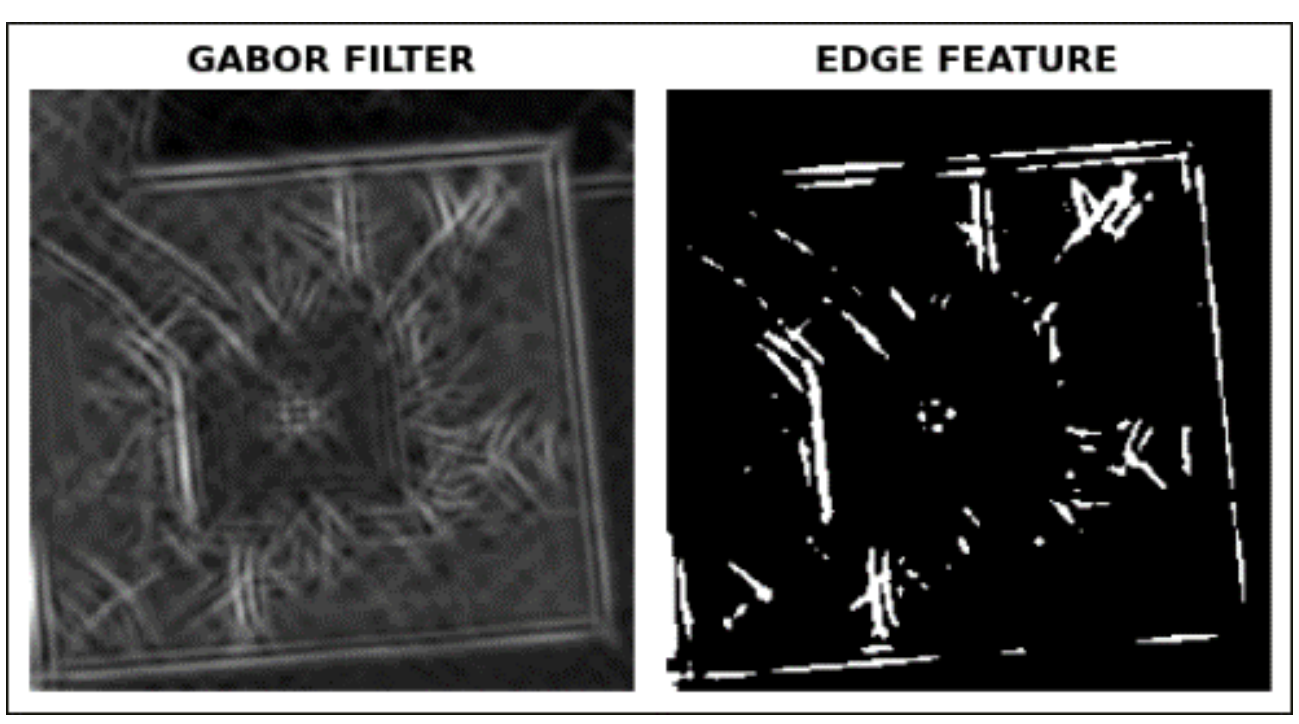

Fig. 9. Feature extraction using Gabor filter and edge detection

Figure 9 decipts that the Gabor filter creates texture and directionality for an image while using edge detection find structural boundaries. By enhancing these features through their respective filters, the model can more easily convey the shape and location of an airplane within the frame.

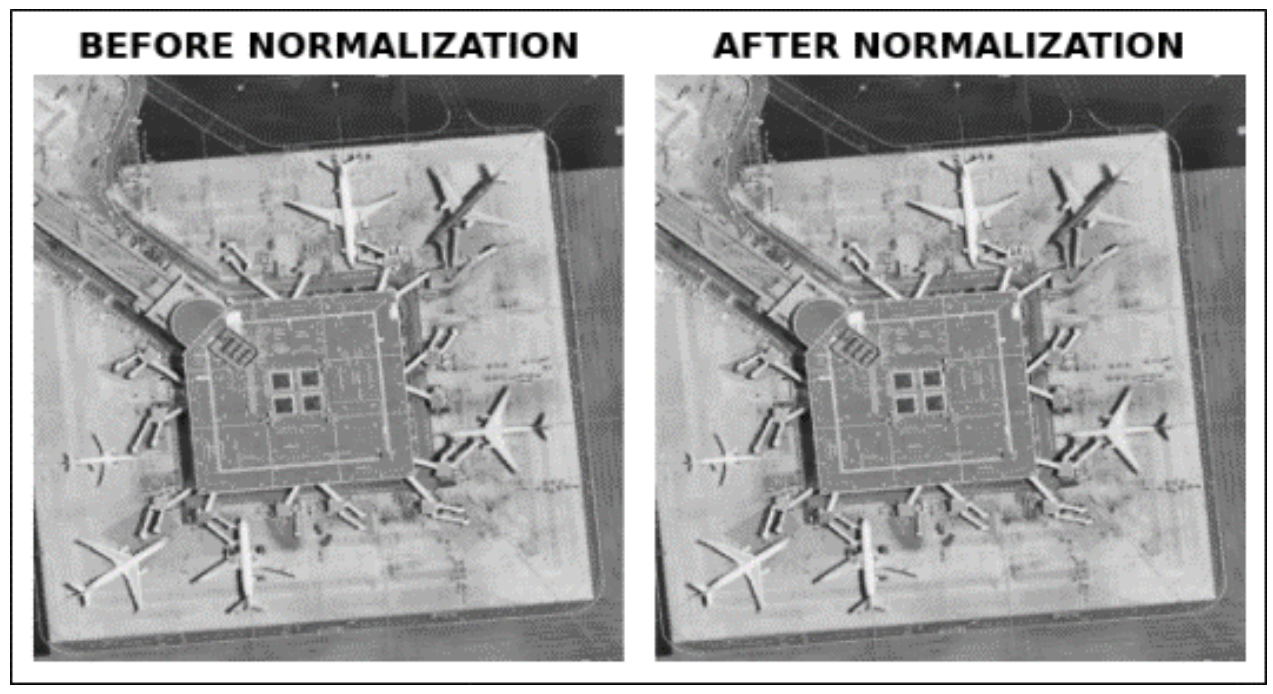

Fig. 10. Image before and after normalization

Figure 10 shows the difference in brightness and contrast between an image before and after normalization. Normalization makes things easier to see and makes sure that the brightness is evenly spread across the whole image.

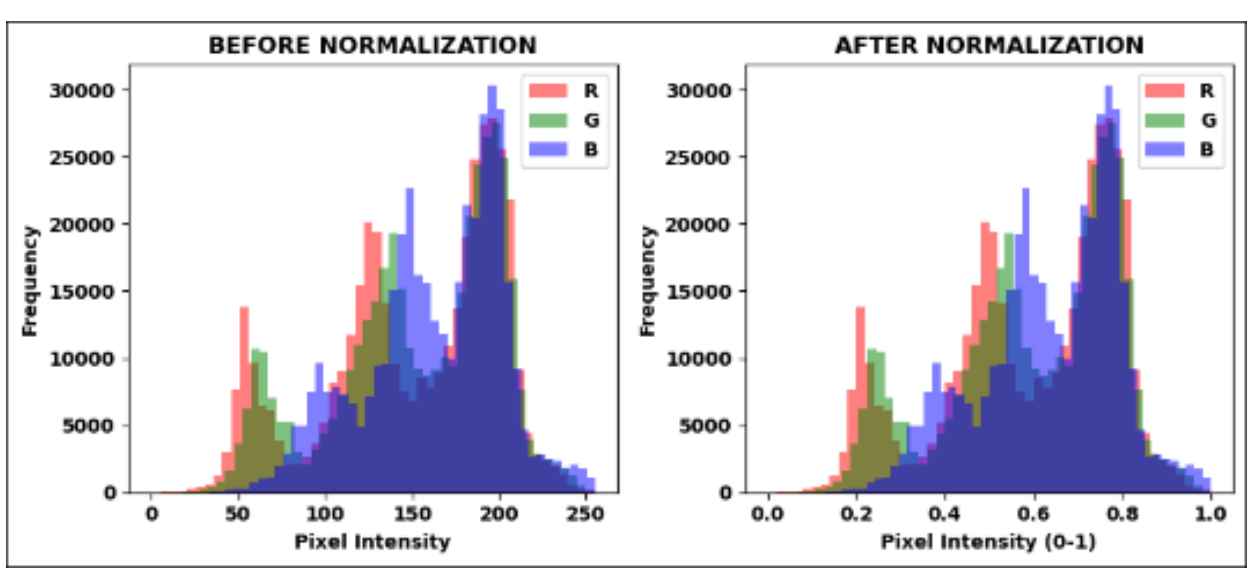

Fig. 11. RGB histogram comparison before and after normalization

Figure 11 indicates RGB channel histograms before and after normalization, which shows that pixel intensity values are moved around. After normalization, the intensity values are adjusted to fit into a single range, which makes the colour channels more consistent.

TABLE II. PERFORMANCE EVALUATION METRICS OF PROPOSED MODEL

| Performance Metrics | |
|---|---|
| **Method** | **Value** |
| mAP@50 | 95% |
| mAP@50 | 70% |
| Precision | 97% |
| Recall | 85% |
| F1-score | 91% |

Table II shows the performance metrics of the proposed model. The mAP@50 value is 95%, which is higher than the 70% value, which means that the model is better at detecting things. The model also shows that it works well by having a precision of 97% and a recall of 85%, which confirms that make accurate predictions.

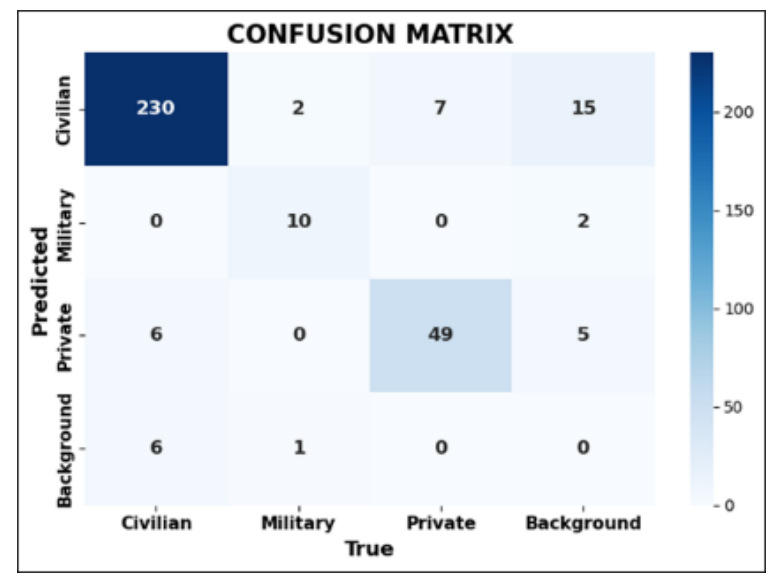

Fig. 12. Confusion matrix for multiclass classification

Figure 12 shows a confusion matrix that shows that the model performed at classifying four groups: Civilian, Military, Private, and Background. High values along the diagonal mean that the predictions were correct, while values that are not on the diagonal mean that the classes were mixed up.

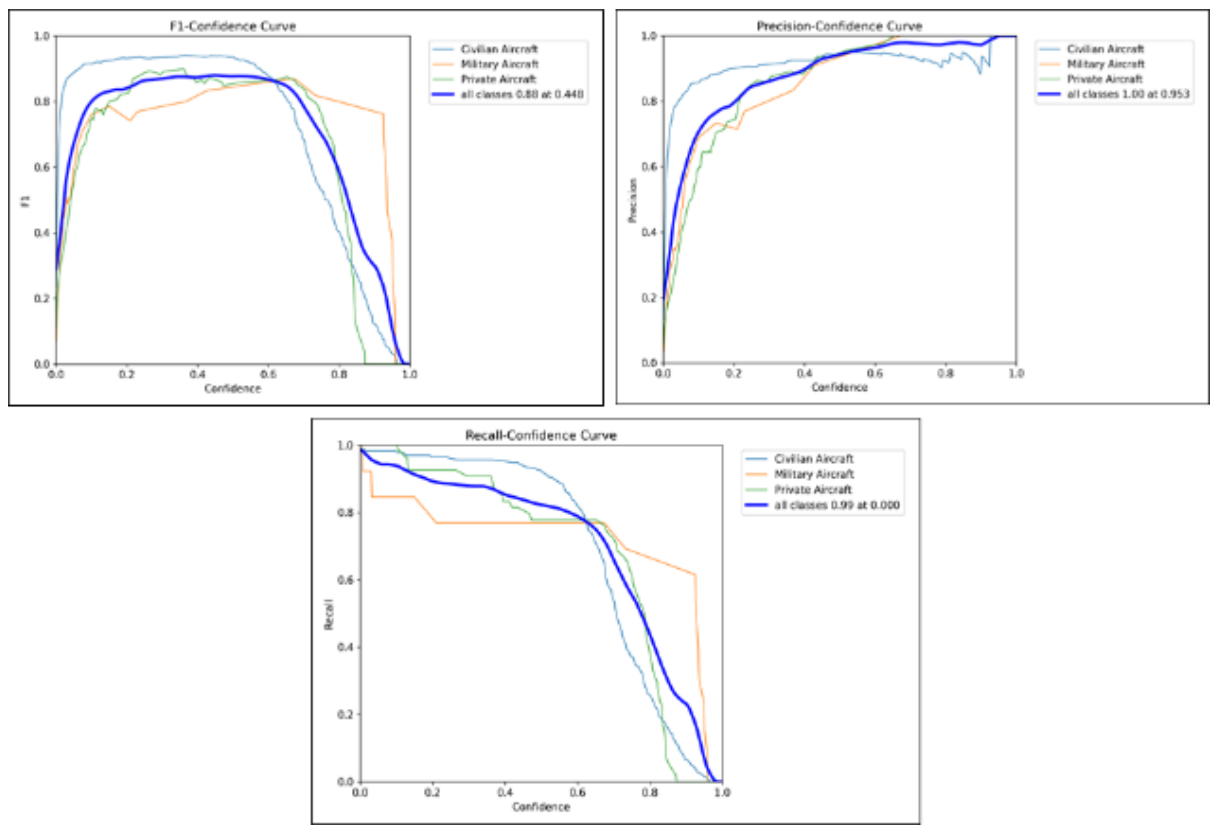

Fig. 13. Confidence based performance curves for aircraft classification

Figure 13 represents that the F1 score, precision, and recall change with different confidence thresholds for different types of aircraft. The curves show the trade-offs in model performance, showing the way confidence levels affect detection accuracy across categories.

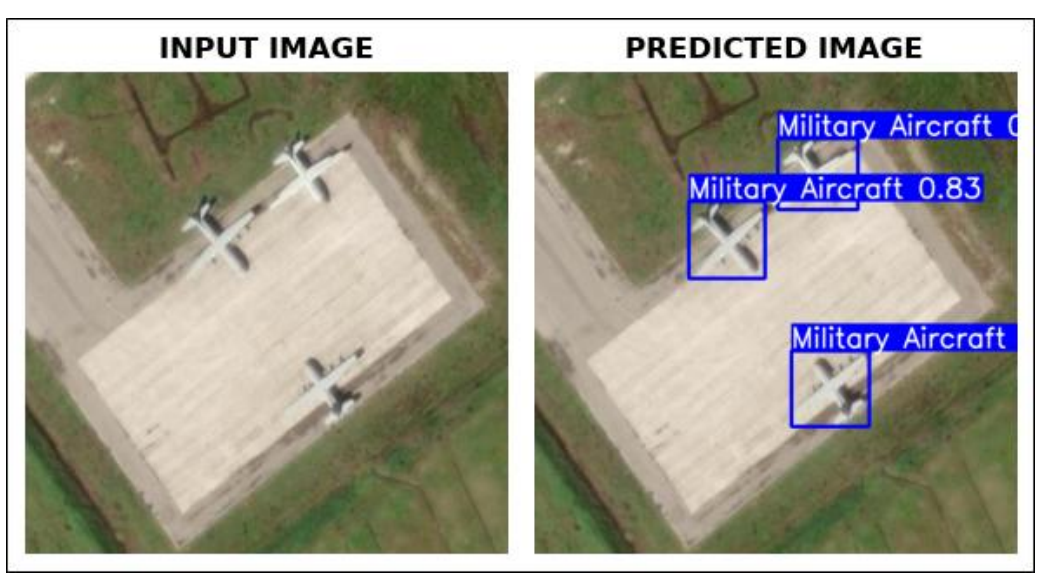

Fig. 14. Aircraft detection using object detection model

Figure 14 shows an aerial view of an airfield with several planes on it as the input. The model output shows detected military aircraft with bounding boxes and confidence scores next to them.

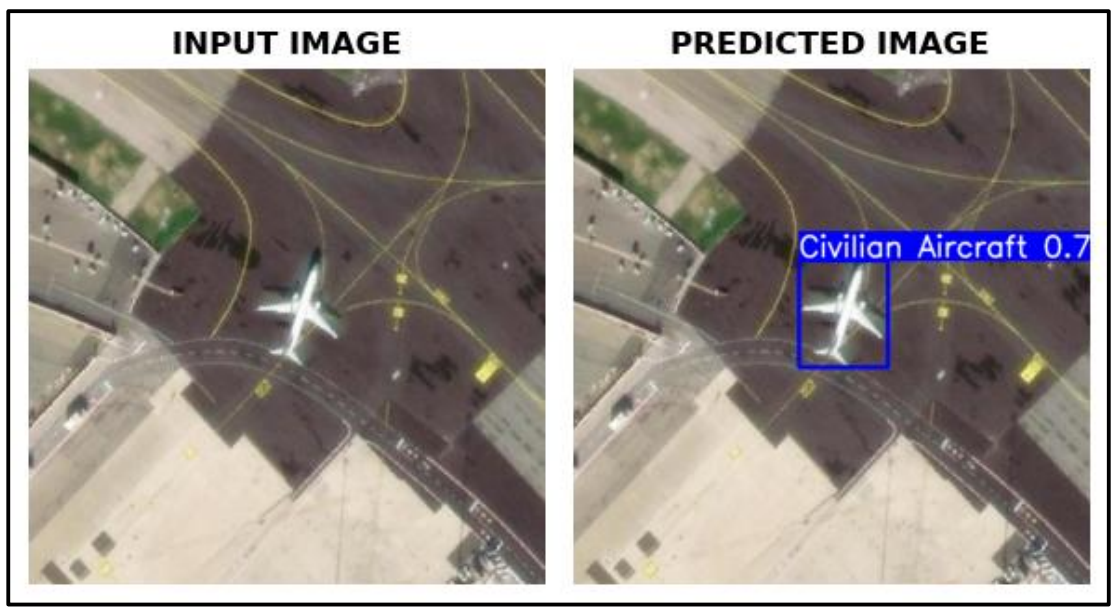

Fig. 15. Civilian aircraft detection using object detection model

Figure 15 shows an aerial view of an airport scene with a civilian plane in it. The model output shows the predicted class and its confidence score, as well as a bounding box around the plane.

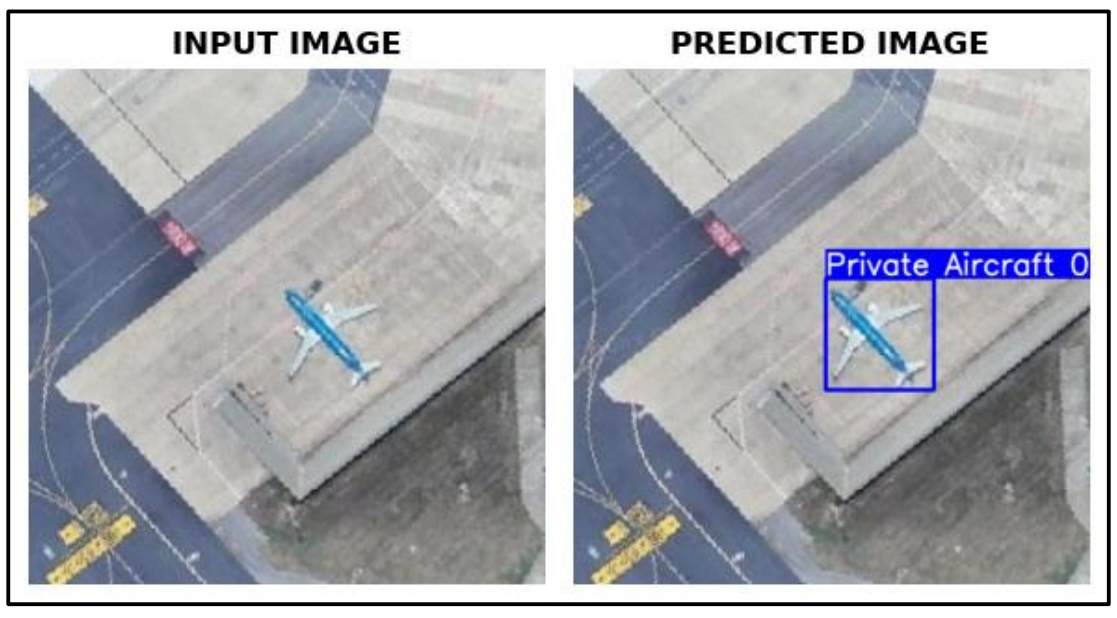

Fig. 16. Aircraft detection using computer vision model

Figure 16 decipts that the model gets an aerial images of an airport apron with a parked plane as input. The model correctly finds and labels the airplane with a bounding box that says "Private Aircraft 0."

TABLE III. COMPARISON OF MAP PERFORMANCE WITH EXISTING METHODS

| mAP | |
|---|---|
| **Method** | **Value** |
| YOLOv5 [13] | 81% |
| YOLOv7 Tiny [14] | 93% |
| Proposed | 95% |

Table III. Represents that the proposed model's mAP compares to that of other models. The proposed model has the highest mAP, which is 95%. It beats YOLOv5 (81%) and YOLOv7 Tiny (93%), showing that it is better at finding things.

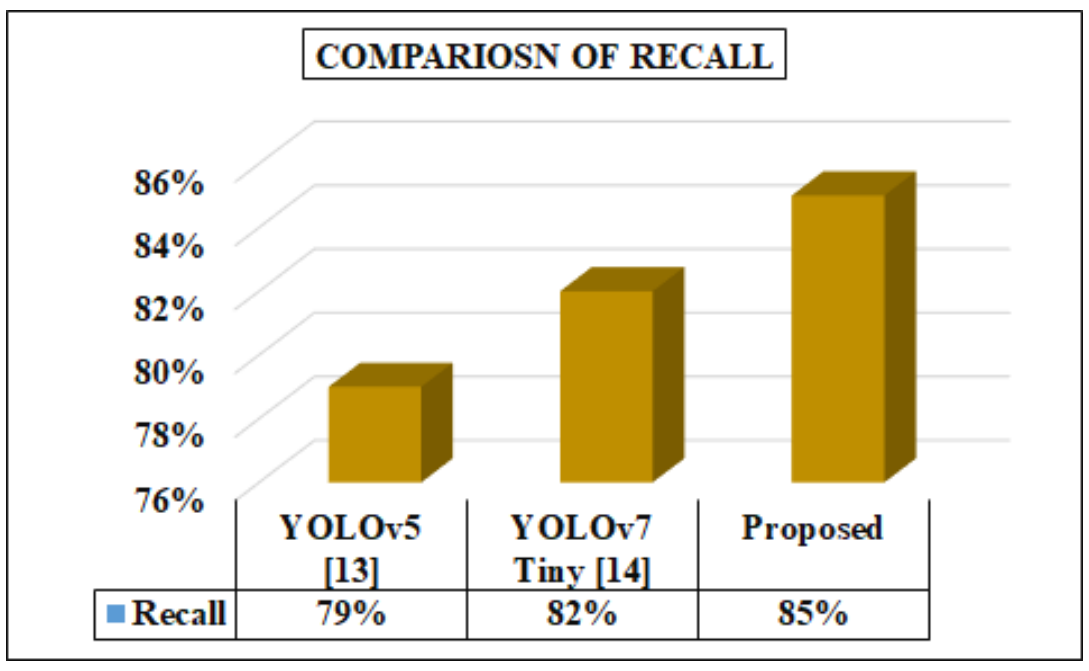

Fig. 17. Comparison of recall performance among detection models

Figure 17 shows the recall values for YOLOv5, YOLOv7 Tiny, and the proposed model. It shows that the detection ability gets better over time. The proposed model has the best recall rate of 85%, which is better than YOLOv7 Tiny (82%) and YOLOv5 (79%).

## V. CONCLUSION

This study presented a DL methodology for aircraft detection in satellite imagery, proficiently managing intricate scenes and variations in aircraft appearance. The proposed model, which is ensured by pre-processing and feature extraction methods, shows better detection ability. It has a high mAP of 95%, a precision of 97%, a recall of 85%, and an F1-score of 91%. This is better than other models like YOLOv5 and YOLOv7 Tiny. The results show that accurate classification and reliable detection are possible for many types of aircraft. The model works well and are able used in real life for applications like surveillance and air traffic monitoring. It could also be improved for use in real time.